\documentclass{article}
\usepackage{acra}
\usepackage{amsmath}
\usepackage{amssymb}
\usepackage{enumitem}
\usepackage{graphicx} 
\usepackage{subcaption} 
\usepackage{booktabs}

\usepackage[ruled,vlined]{algorithm2e}
\usepackage{algpseudocode}
\usepackage{tikz}
\usepackage{svg} 
\usepackage{import}
\usetikzlibrary{arrows.meta,positioning,shapes.geometric,fit,calc}
\usepackage{dblfloatfix}

\title{%
\vspace*{-2.0em}

{\raggedright
\small
Cite as: Shuaidong Ji, Mahdi Bamdad, and Francisco Cruz.
SWIFT-Nav: Stability-Aware Waypoint-Level TD3 with Fuzzy Arbitration for UAV Navigation in Cluttered Environments.
\textit{Proceedings of the Australasian Conference on Robotics and Automation (ACRA), 2025.}
\par}

\vspace{0.45em}
\hrule
\vspace{0.9em}

\Large
SWIFT-Nav: Stability-Aware Waypoint-Level TD3 with Fuzzy Arbitration for UAV Navigation in Cluttered Environments
}

\author{%
  Shuaidong Ji$^{1}$ \quad
  Mahdi Bamdad$^{1}$ \quad
  Francisco Cruz$^{1,2}$ \\[3pt]
  $^{1}$School of Computer Science and Engineering, University of New South Wales, Sydney, Australia\\
  $^{2}$Escuela de Ingeniería, Universidad Central de Chile, Santiago, Chile\\[3pt]
  \texttt{joffrey.ji@unsw.edu.au, m.bamdad@unsw.edu.au, f.cruz@unsw.edu.au}\\[4pt]
}

\usepackage{titlesec}
\titlespacing*{\paragraph}{0pt}{0.3ex plus .2ex minus .2ex}{1em}

\begin{document}
\maketitle
\vspace{0.6em}

\begin{abstract}
Efficient and reliable UAV navigation in cluttered and dynamic environments remains challenging. We propose \textbf{SWIFT-Nav}---\emph{Stability-aware Waypoint-level Integration of Fuzzy arbitration and TD3 for Navigation}---a TD3-based navigation framework that achieves fast, stable convergence to obstacle-aware paths. The system couples a sensor-driven perception front end with a TD3 waypoint policy: the perception module converts LiDAR ranges into a confidence-weighted safety map and goal cues, while the TD3 policy is trained with Prioritized Experience Replay to focus on high-error transitions and a decaying $\varepsilon$-greedy exploration schedule that gradually shifts from exploration to exploitation. A lightweight fuzzy-logic layer computes a safety score from radial measurements and, near obstacles, gates mode switching and clamps unsafe actions; in parallel, task-aligned reward shaping—combining goal progress, clearance, and switch-economy terms—provides dense, well-scaled feedback that accelerates learning. Implemented in Webots with proximity-based collision checking, our approach consistently outperforms baselines in trajectory smoothness and generalization to unseen layouts, while preserving real-time responsiveness. These results show that combining TD3 with replay prioritization, calibrated exploration, and fuzzy-safety rules yields a robust and deployable solution for UAV navigation in cluttered scenes.
\end{abstract}

\section{Introduction}
Unmanned Aerial Vehicles (UAVs) have moved from prototypes to operational assets in disaster response, environmental monitoring, infrastructure inspection, and autonomous logistics ~\cite{mu22}. As expectations shift from remote teleoperation to autonomy on board, UAVs must navigate partially known cluttered spaces with minimal human intervention ~\cite{yin24}. In urban canyons, forests, and post-disaster zones, this demands real-time perception of hazards, continual replanning, and safe execution under sensing noise and state uncertainty ~\cite{fag24}.

Model-based planning and classical control work well when maps are accurate and conditions are benign; however, in complex, cluttered, and partially observed environments their assumptions break down, and performance degrades. Deep reinforcement learning (DRL) learns adaptive behaviors directly from interaction with the environment~\cite{zha24}. Within DRL, Twin Delayed Deep Deterministic Policy Gradient (TD3) has strong results in continuous control, yet practical UAV navigation still faces key hurdles: rewards are often sparse and delayed, exploration can be unstable in safety-critical settings, and the system must operate under tight on-board compute and latency constraints~\cite{hus24}.

This work develops a TD3-based navigation framework targeted at cluttered environments and real-time deployment. The core idea is to learn at the path-planning level while delegating low-level execution to a lightweight line-following controller, thereby stabilizing the closed loop and reducing sensitivity to actuator latency. Reinforcement learning is engaged only near hazards; elsewhere a compute-free nominal mode progresses toward the goal. Stability-aware arbitration (hysteresis, debounce, and minimum dwell time) and a fuzzy safety prior suppress spurious switches. To improve sample efficiency and policy robustness, we combine prioritized experience replay (PER)~\cite{guo24} with a decaying $\varepsilon$-greedy schedule and task-aligned reward shaping ~\cite{millan19}. A lightweight checker guards against degenerate proposals and accelerates recovery in ambiguous situations. The full system is implemented in Webots, enabling controlled studies with realistic vehicle dynamics.

This paper makes four contributions: (1) a hybrid TD3 navigation architecture that plans at the waypoint level and executes via a shared line-following controller, enabling fast and stable closed-loop flight; (2) a stability-aware arbitration mechanism with fuzzy safety scoring that invokes RL only when necessary and suppresses unnecessary or rapid back-and-forth mode switching; (3) a sample-efficient training pipeline that combines PER, a decaying $\varepsilon$-greedy policy, and task-aligned reward shaping, together with a lightweight checker to regularize updates; and (4) a Webots-based simulation pipeline for Apple Silicon (M-series) Macs, filling a tooling gap: unlike mainstream simulators such as AirSim that target Windows/Linux ~\cite{ko25} and lack native Apple-Silicon support, our setup runs natively on recent macOS while preserving real-time responsiveness.

\section{Related Work}

\noindent
Research on learning-based UAV navigation in cluttered scenes has evolved along two interacting
axes: \emph{(i) how supervision signals are formed} from reward shaping/curricula to the way
range data are converted into sector-wise risk scores and \emph{(ii) how the policy is optimized
and embedded} from TD3 variants that stabilize updates to hybrid stacks that arbitrate between
local RL and classical controllers. We organize the discussion accordingly: first, densified
rewards and their limitations; second, optimization-oriented extensions of TD3; third, common
sector-average LiDAR front ends; and finally, hybrid navigation with mode-switch logic. This
structure sets up the gaps we highlight next and motivates our stability-aware, confidence-modulated approach.

\paragraph{Reward shaping and curricula.}
A common remedy for sparse rewards is to combine a terminal goal bonus with a dense, progress-based step signal typically the reduction in goal distance, $d(s_{t-1})-d(s_t)$, minus a fixed time cost $c$, often after a demonstration- or curriculum-style warm start \cite{he20}. Such shaping suppresses early catastrophic exploration, but the hand-tuned weights (progress, smoothness, collision, etc.) are tightly coupled to a particular map topology and sensing stack. This entanglement distorts credit assignment and weakens transfer to environments with different obstacle densities and dynamics; many implementations also terminate on crash without additional penalties, further biasing the learning signal.

\paragraph{TD3 variants for stability/efficiency.}
A parallel thread improves \emph{optimization} while keeping the perception–action interface unchanged. Typical ingredients include \emph{Average-TD3} (using the mean of twin critics in the target to curb underestimation), \emph{prioritized replay} to focus updates on informative transitions, and light regularizers such as \emph{target-policy smoothing}, \emph{delayed actor updates}, and \emph{decaying exploration noise} \cite{Luo24,Li24}. These tweaks generally reduce variance and accelerate convergence in cluttered scenes, but they do not address how perception encodes risk or how control authority is switched.

\paragraph{Sector–average perception front ends.}
Following ~\cite{Li24}, many systems aggregate LiDAR in a forward ${\sim}180^\circ$ fan and assign \emph{uniform confidence} to rays/sectors, scoring each sector by the \emph{average} of distance memberships over its $N$ rays. While simple and stable, this uniform treatment ignores corridor width and directional reliability: a few short ranges can depress the sector score for an otherwise traversable direction, so partially free sectors are penalized almost as much as blocked ones. The planner thus tends to conservative detours, skirting obstacle clusters instead of safely threading narrow gaps, yielding longer paths on tight maps despite preserved safety.

\paragraph{Hybrid navigation and mode arbitration.}
Global–local hybrids (global planner + local RL) are common, yet authority transfer is often triggered by ad-hoc thresholds or instantaneous risk, without stability guarantees. Prior work seldom models \emph{hysteresis}, \emph{minimum dwell time}, or \emph{debounce} to prevent switch chatter, nor does it fold sensing delay and estimator drift into the arbitration logic. Reward designs also rarely include an explicit \emph{switch-economy} term alongside progress and clearance.

\paragraph{Hazard-Triggered Waypoint Control.}
Closest to our setup is \cite{Liu24}, who explored 3D autonomous navigation using an option-critic framework with two high-level behaviors (“look-around” and “frontier navigation”) guided by classical planning. Their goal was efficient coverage mapping, whereas our work addresses point-to-point UAV navigation in cluttered spaces through a stability-aware hybrid framework. Specifically, a TD3 waypoint policy activates only near hazards, governed by a fuzzy safety arbiter and stabilized through hysteresis, debounce, and dwell-time constraints, while a trajectory checker ensures safe waypoint execution. Unlike Liu et al., who evaluate exploration coverage, we emphasize path stability and efficiency—achieving smoother, non-oscillatory trajectories quantified through ablation on stability and safety modules.

\paragraph{Gaps and our angle.}
Despite notable advances, three gaps persist: (i) missing confidence-aware, directional reasoning for valuing candidate directions; (ii) limited treatment of latency and mode-switch stability in hybrids; and (iii) rewards that omit pressure toward economical switching. Our work targets these gaps via a stability-aware hybrid architecture, confidence-modulated (fuzzy) safety logic, and sample-efficient TD3 (PER with calibrated exploration), yielding faster and more reliable navigation in cluttered environments.

~\section{Methodology}
We decouple high-level decision making from low-level actuation in SWIFT-Nav (Stability-Aware Waypoint-Level TD3 with Fuzzy Arbitration). A TD3 agent outputs short-horizon waypoints (trajectory segments), and a lightweight motion controller tracks these waypoints while handling stabilization and motor commands. Training and inference therefore occur at the waypoint interface, and the controller executes the resulting path with standard tracking laws and safety limits.

To assemble the full SWIFT-Nav stack, we add a perception-guided fuzzy layer used during early operation, task-aligned reward terms (progress, clearance, and switch economy), and a perception-sensitivity schedule that adapts to local obstacle density. The state encoding concatenates local obstacle ranges, global goal cues, and directional scans into a compact vector. Training uses Prioritized Experience Replay, and execution alternates between a nominal 'Travel' mode and an RL mode via a finite-state policy switch.

\subsection{System Structure}
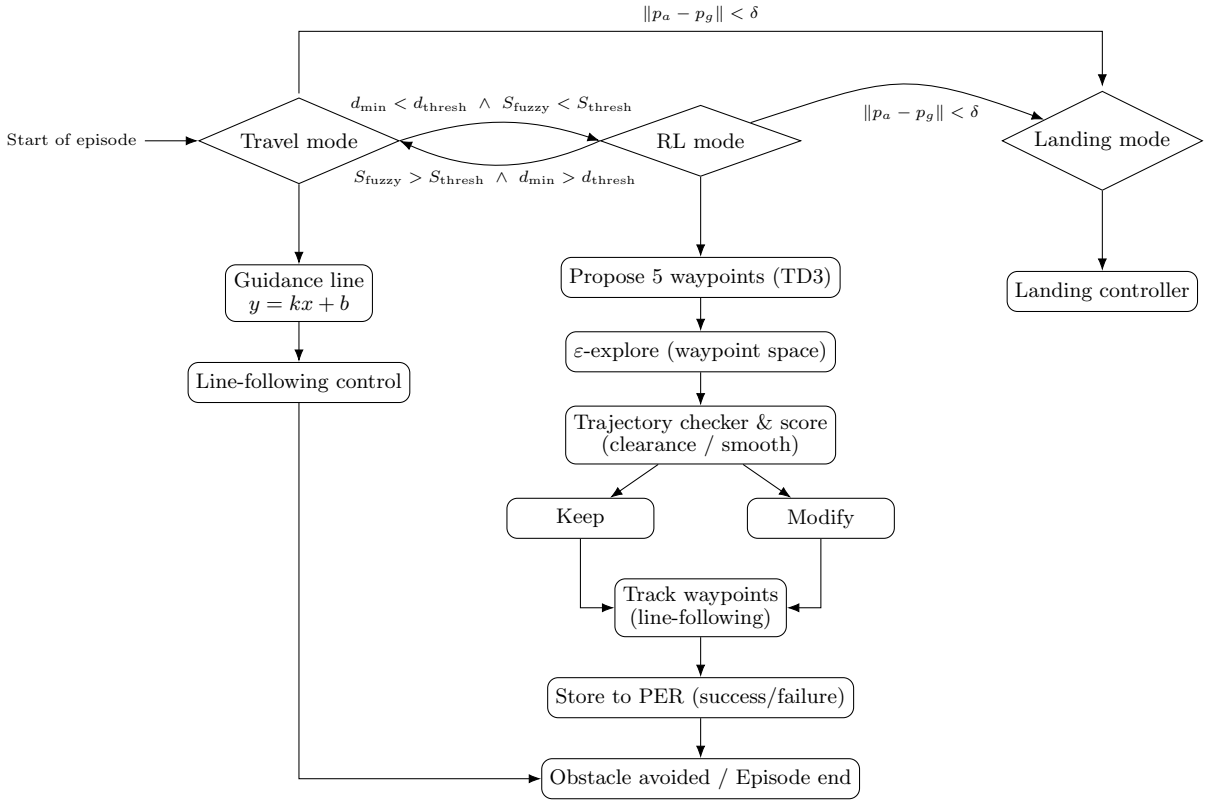
\begin{figure*}[!t]
\centering
\begingroup
\tikzset{every picture/.style={scale=0.9, transform shape}}
\resizebox{0.90\textwidth}{!}{%
\begin{tikzpicture}[
  node distance=6mm and 8mm,
  >=Latex, font=\small,
  box/.style={rectangle, rounded corners, draw, align=center,
              minimum width=22mm, minimum height=6mm},
  decision/.style={diamond, draw, aspect=2, align=center,
                   inner ysep=1pt, minimum width=30mm},
  line/.style={-Latex}
]

\node[decision]                        (normal) {Travel mode};
\node[decision, right=30mm of normal]  (rl)     {RL mode};
\node[decision, right=30mm of rl]      (land)   {Landing mode};

\coordinate (startpt) at ([xshift=-8mm]normal.west); 
\draw[line] (startpt) -- (normal.west);             
\node[anchor=east, font=\scriptsize] at (startpt) {Start of episode};

\draw[line] (normal.east) to[bend left=18]
  node[pos=0.37, above=0.8ex, xshift=2.5mm,
       font=\scriptsize, fill=white, inner sep=1pt]
  {$d_{\min}<d_{\text{thresh}}\ \land\ S_{\text{fuzzy}}<S_{\text{thresh}}$}
  (rl.west);

\draw[line] (rl.west) to[bend left=24]
  node[pos=0.73, below=0.8ex, xshift=6mm,
       font=\scriptsize, fill=white, inner sep=1pt]
  {$S_{\text{fuzzy}}>S_{\text{thresh}}\ \land\ d_{\min}>d_{\text{thresh}}$}
  (normal.east);

\coordinate (topMid) at ([yshift=11mm]rl.north);
\draw[line, shorten >=1.5pt, shorten <=1.5pt]
  (normal.north) -- (normal.north |- topMid)
  -- node[above=0.6ex, font=\scriptsize, fill=white, inner sep=1pt]
     {$\lVert p_a - p_g\rVert < \delta$}
  (land.north |- topMid) -- (land.north);

\draw[line] (rl.20) to[out=18,in=160, looseness=1.3]
  node[pos=0.6, below=1.6ex, font=\scriptsize, fill=white, inner sep=1pt]
  {$\lVert p_a-p_g\rVert<\delta$} (land.160);

\node[box, below=12mm of normal] (guide) {Guidance line \\ $y=kx+b$};
\node[box, below=6mm  of guide]  (lf1)   {Line-following control};
\draw[line] (normal) -- (guide);
\draw[line] (guide)  -- (lf1);

\node[box, below=12mm of rl] (prop) {Propose 5 waypoints (TD3)};
\node[box, below=5mm  of prop] (eps) {$\varepsilon$-explore (waypoint space)};
\node[box, below=5mm  of eps]  (chk) {Trajectory checker \& score\\ (clearance / smooth)}
;
\draw[line] (rl) -- (prop);
\draw[line] (prop) -- (eps);
\draw[line] (eps)  -- (chk);

\node[box, below=5mm of chk, xshift=-18mm] (keep) {Keep};
\node[box, below=5mm of chk, xshift= 18mm] (mod)  {Modify};
\draw[line] (chk) -- (keep);
\draw[line] (chk) -- (mod);

\coordinate (midkm) at ($(keep)!0.5!(mod)$);
\node[box, below=9mm of midkm] (track) {Track waypoints\\ (line-following)};
\draw[line] (keep) |- (track);
\draw[line] (mod)  |- (track);

\node[box, below=6mm of track] (per) {Store to PER (success/failure)};
\draw[line] (track) -- (per);

\node[box, below=12mm of land] (landctrl) {Landing controller};
\draw[line] (land) -- (landctrl);

\node[box, below=6mm of per] (done) {Obstacle avoided / Episode end};
\draw[line] (per) -- (done);
\draw[line] (lf1) |- (done);

\end{tikzpicture}}
\endgroup

\caption{SWIFT-Nav overview with abstracted steps and explicit mode transitions.
Flow \emph{starts at} \textbf{Travel mode} (arrow) where the UAV follows a guidance line. 
When the risk trigger fires, control switches to \textbf{RL mode} to propose/score waypoints; either mode transitions to \textbf{Landing mode} once the goal tolerance is met, after which the episode ends.}
\label{fig:swiftnav_overview}
\end{figure*}

Figure~\ref{fig:swiftnav_overview} summarizes SWIFT-Nav. The system runs a three-mode finite-state machine with explicit arbitration: \textit{Travel mode} follows a geometric guidance line ($y=kx+b$); \textit{RL mode} is engaged only near hazards according to a joint distance–safety trigger; and \textit{Landing mode} is entered when the goal tolerance is met. Mode switches are stabilized by hysteresis, debounce, and a minimum dwell time; returning from RL to Travel requires both recovery of the fuzzy safety score and line-of-sight to the goal. Transitions produced in RL are stored in a prioritized replay buffer for sample-efficient learning.

\paragraph{Stability rationale.}
We require asymmetric thresholds (enter vs. exit), a debounce counter (the desired mode must persist for $N$ consecutive steps), and minimum dwell time in the current mode. Exiting RL additionally requires a line-of-sight (LOS) to the goal and a fuzzy safety score above a threshold. Together these rules bound the switching frequency by construction: a new switch cannot occur before the dwell time elapses, and transient risk fluctuations are filtered by the debounce. The LOS guard prevents premature exits that would immediately re-enter RL due to occlusions. In Section 5 we show this yields the characteristic pattern of steps ↓ / reward ↑ with few, non-oscillatory switches.

\subsection{Travel mode}
\label{subsec:travel}

\textit{Travel mode} provides a light–weight, deterministic backbone that moves the UAV toward the goal without invoking RL. Upon \emph{entering} this mode we instantiate a \emph{guidance line} through the current UAV position and the goal; the line remains \emph{frozen} while the system stays in Travel mode and is only recomputed on \emph{mode entry} (or if the goal changes). This makes the line ``dynamic across modes'' (updated at each Travel–entry) rather than continuously drifting every time step, which prevents oscillatory behavior near the switching boundaries.

\paragraph{Line generation.}
Let $\mathbf{p}_d=(x_d,y_d)$ be the UAV position at the moment we enter Travel mode and $\mathbf{p}_g=(x_g,y_g)$ be the goal in the global $(x,y)$ plane. Define
\begin{equation}
\label{eq:travel_defs}
\Delta x = x_g - x_d,\qquad
\Delta y = y_g - y_d,\qquad
\psi_g = \operatorname{atan2}(\Delta y,\Delta x).
\end{equation}
When $|\Delta x|>\varepsilon$, we use the explicit form
\begin{equation}
\label{eq:line_kb}
y = kx + b,\qquad
k = \frac{\Delta y}{\Delta x},\qquad
b = y_d - k\,x_d .
\end{equation}
Otherwise the path can be treated as a vertical line ($x=x_d$). For controller implementation it is convenient to also keep the \emph{parametric} line and its unit travel.
\begin{equation}
\label{eq:param_line}
\mathbf{p}(s)=\mathbf{p}_d+s\,\mathbf{t},\qquad
\mathbf{t}=\begin{bmatrix}\cos\psi_g\\ \sin\psi_g\end{bmatrix},\qquad
\mathbf{n}=\begin{bmatrix}-\sin\psi_g\\ \cos\psi_g\end{bmatrix}.
\end{equation}

\paragraph{Errors and control.}
Given the current pose $(\mathbf{p},\psi)$, the signed cross–track and heading errors are
$e_\perp=\mathbf{n}^\top(\mathbf{p}-\mathbf{p}_d)$ and $e_\psi=\operatorname{wrap}(\psi-\psi_g)$.
A simple line–following law then commands turning (or yaw–rate) using a Stanley/PID hybrid:
\begin{equation}
\label{eq:stanley}
\delta = e_\psi + \arctan\!\left(\frac{k_s\,e_\perp}{v+\epsilon}\right),
\end{equation}
or equivalently a PD form with saturations,
$\omega = k_\psi\,e_\psi + k_y\,e_\perp$.
Progress along the line is measured by
$s=\mathbf{t}^\top(\mathbf{p}-\mathbf{p}_d)$.
Once $s$ exceeds the remaining line length, or the UAV enters the landing tolerance,
arbitration transitions to \emph{Landing mode}.
Disturbances causing lateral drift are corrected via $e_\perp$ rather than re-fitting the line,
preserving stability.

\subsection{RL mode}
\paragraph{Engagement trigger.}
RL is engaged only when both (i) the minimum range to surrounding obstacles
drops below a threshold and (ii) the fuzzy safety score indicates risk:
\begin{equation}
d_{\min} < d_{\text{thresh}} \;\land\; S_{\text{fuzzy}} < S_{\text{thresh}} .
\label{eq:rl_trigger}
\end{equation}
Let $\mathcal{O}=\{\mathbf{p}_i\}$ denote the set of obstacle points (or obstacle
representatives) with $\mathbf{p}_i\in\mathbb{R}^{2}$ the position of the $i$-th
obstacle, and let $\mathbf{p}_d$ be the drone position. The minimum range is
\begin{equation}
d_{\min}\;=\;\min_{\;i:\ \|\mathbf{p}_i-\mathbf{p}_d\|_2\le R}\ \|\mathbf{p}_i-\mathbf{p}_d\|_2,
\label{eq:dmin}
\end{equation}
with sensing radius $R$.

\begin{figure}[t]
  \centering
  \includegraphics[width=1\linewidth]{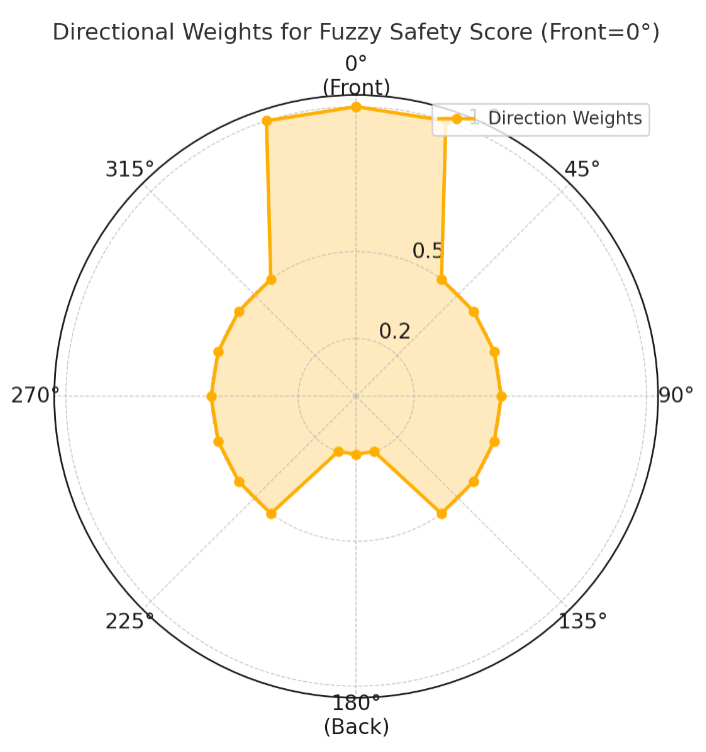}
  \caption{Directional weights for the fuzzy safety score. 
  Front sectors ($0^\circ$–$30^\circ$ and $330^\circ$–$360^\circ$): $w{=}1.0$;
  side sectors ($30^\circ$–$150^\circ$, $210^\circ$–$330^\circ$): $w{=}0.5$;
  rear ($150^\circ$–$210^\circ$): $w{=}0.2$. Higher weight = higher directional importance.}
  \label{fig:fuzzy_weight}
\end{figure}

\paragraph{Fuzzy safety score.}
We cast a set of radial directions $\{\theta_i\}$ around the platform and map the measured
range $d_i$ along direction $\theta_i$ to a distance–safety membership via a linear S‐shaped
rule
\begin{equation}
r(d_i)\;=\;
\begin{cases}
1, & d\ge 10,\\[2pt]
0, & d\le 2,\\[2pt]
(d_i-2)/8, & \text{otherwise,}
\end{cases}
\label{eq:membership}
\end{equation}
and encode directional importance with weights $w_i$ (shown as Figure ~\ref{fig:fuzzy_weight}):
\begin{equation}
w_i\;=\;
\begin{cases}
1.0, & \theta_i\in[330^\circ,360^\circ]\ \cup\ [0^\circ,30^\circ]\quad\text{(front)},\\[2pt]
0.5, & \theta_i\in(30^\circ,150^\circ)\ \cup\ (210^\circ,330^\circ)\quad\text{(sides)},\\[2pt]
0.2, & \theta_i\in[150^\circ,210^\circ]\quad\text{(rear)}.
\end{cases}
\label{eq:weights}
\end{equation}
The confidence–aware score is the weighted average
\begin{equation}
S_{\text{fuzzy}}\;=\;
\frac{\sum_i r(d_i)\,w_i}{\sum_i w_i}\ \in[0,1],
\label{eq:s_fuzzy}
\end{equation}
which biases decisions toward frontal clearance while avoiding over-penalizing partially free sectors.

\paragraph{TD3 backbone.}
We adopt Twin Delayed DDPG (TD3) with clipped double-$Q$, delayed actor updates, and target
policy smoothing. The critic target is
\begin{equation}
\label{eq:td3_target}
\begin{aligned}
y_t &= r_t + \gamma \min_{i\in\{1,2\}} Q_{\theta_i'}\!\left(s_{t+1},\,\tilde a_t\right),\\
\tilde a_t &= C_{\phi'}(s_{t+1}) + \varepsilon',\quad
\varepsilon' \sim \mathrm{clip}\big(\mathcal N(0,\sigma),-c,c\big),
\end{aligned}
\end{equation}
and targets are softly updated as
\begin{equation}
\theta_i' \leftarrow \tau \theta_i + (1-\tau)\theta_i',
\qquad
\phi' \leftarrow \tau \phi + (1-\tau)\phi'.
\label{eq:td3_soft}
\end{equation}

\paragraph{Waypoint proposal and selection (geometric $\varepsilon$-exploration).}
In RL mode , the actor proposes five
intermediate waypoints $\mathcal{W}=\{\mathbf{w}_k\}_{k=1}^5$.
Instead of the common \emph{action-space} exploration that directly perturbs low-level
controls (e.g., left/right or yaw-rate noise), we perform \emph{geometry-space}
exploration by adding small, bounded offsets to the waypoint coordinates:
\begin{equation}
\label{eq:geom_eps}
\begin{aligned}
\tilde{\mathbf{w}}_k &= \mathbf{w}_k + \varepsilon_t\,\Delta\mathbf{w}_k,\\
\varepsilon_t &= \varepsilon_0 \lambda^{t}, \qquad 0<\lambda<1,\qquad \|\Delta\mathbf{w}_k\| \le \rho .
\end{aligned}
\end{equation}
which preserves path continuity and searches locally around the five nominal points; as
training proceeds, $\varepsilon_t$ decays to transition from broad search to exploitation.

\paragraph{Trajectory Checker \& score}
A lightweight \textit{Trajectory Checker and score} evaluates the candidate polyline
$\tilde{\tau}=(\tilde{\mathbf{w}}_1,\dots,\tilde{\mathbf{w}}_5)$ for (i) collision along
the piecewise-linear path, (ii) minimum clearance margin, (iii) curvature/smoothness, and
(iv) progress toward the goal. If the current nominal polyline violates clearance, the
checker applies constrained corrective displacements only to the offending subset of
waypoints (modify) to satisfy the margin; otherwise the original proposal is kept:
\begin{equation}
\tau^\star=\arg\max_{\tau\in\{\text{keep},\,\text{modify}\}}\ \mathrm{Score}(\tau).
\label{eq:checker_score}
\end{equation}
The selected polyline is then executed by the line-following controller.

\paragraph{Priority Experience Reply (PER).}
Transitions are stored in a Prioritized Experience Replay buffer and sampled with
\begin{equation}
P(i)=\frac{p_i^\alpha}{\sum_k p_k^\alpha},\qquad 
w_i=\Big(\frac{1}{N\,P(i)}\Big)^\beta,
\label{eq:per}
\end{equation}
followed by IS re-weighting within the minibatch; priorities are refreshed from TD errors.
We follow the abstract procedure in Algorithm~\ref{alg:td3_per} (PER with TD3).
\begin{algorithm}[t]
\caption{TD3 with Prioritized Experience Replay (abstract view)}
\label{alg:td3_per}
\DontPrintSemicolon
\SetKwInOut{Input}{Inputs}
\SetKwInOut{Output}{Output}

\Input{Actor $\pi_\phi$, critics $Q_{\theta_1},Q_{\theta_2}$ and target nets; replay buffer $\mathcal D$ with priorities; hyperparameters $(\gamma,\tau,d,\alpha,\beta_0\!\to\!\beta_1)$}
\Output{Updated $\pi_\phi, Q_{\theta_1}, Q_{\theta_2}$}

\BlankLine
\textbf{Init:} set $p_{\max}\!\leftarrow\!1$; initialize networks and targets.\;

\For{each environment step}{
  \textbf{Interact}: select action from actor with exploration; observe transition $(s,a,r,s',\mathrm{done})$.\;
  \textbf{Store}: push transition into $\mathcal D$ with priority $p_{\max}$.\;

  \If{\textbf{update step}}{
    \textbf{PER sample}: $(\mathcal B, w)\leftarrow\textsc{SamplePER}(\mathcal D;\alpha,\beta)$ \tcp*{returns indices and IS weights}
    \textbf{Targets}: $y\leftarrow\textsc{ComputeTargetsTD3}(\mathcal B)$ \tcp*{policy smoothing + twin critics}
    \textbf{Critic update}: \textsc{UpdateCritics}$(Q_{\theta_1},Q_{\theta_2};\mathcal B,y,w)$.\;

    \If{\textbf{delayed step} ($\bmod\,d=0$)}{
      \textbf{Actor update}: \textsc{UpdateActor}$(\pi_\phi;Q_{\theta_1})$.\;
      \textbf{Soft target update}: \textsc{SoftUpdateTargets}$(\tau)$.\;
    }

    \textbf{Priority write-back}: $\delta\leftarrow\textsc{TDerror}(Q_{\theta_1};\mathcal B,y)$;\;
    \textsc{UpdatePriorities}$(\mathcal D,\mathcal B,|\delta|)$; $p_{\max}\!\leftarrow\!\max(|\delta|)$.\;
    \textbf{Anneal IS exponent}: $\beta\leftarrow\textsc{Anneal}(\beta,\beta_1)$.\;
  }
}
\end{algorithm}

\subsection{Landing mode}

\paragraph{Trigger and safety (compact).}
\begin{subequations}\label{eq:land_group}
\begin{align}
\|\mathbf{p}-\mathbf{p}_g\| &< \delta_{\text{in}}, \label{eq:land_trigger}\\
S_{\text{fuzzy}} &> S_{\text{land}}, \qquad d_{\min} > d_{\text{land}}. \label{eq:land_safety}
\end{align}
\end{subequations}
Landing exits only if $\|\mathbf{p}-\mathbf{p}_g\|>\delta_{\text{out}}$ with
$\delta_{\text{out}}>\delta_{\text{in}}$ (hysteresis).

\paragraph{Final–approach controller.}
Let \(\mathbf{e}_p=\mathbf{p}_g-\mathbf{p}\) and
\(\hat{\mathbf{t}}=\mathbf{e}_p/\|\mathbf{e}_p\|\) be the approach direction.
We align yaw to \(\hat{\mathbf{t}}\) and regulate forward speed with a distance-based
profile
\begin{equation}
\label{eq:land_speed}
v_{\parallel} \;=\; 
\mathrm{clip}\!\big(k_v\|\mathbf{e}_p\|,\ v_{\min},\ v_{\max}\big),\qquad
\dot{\mathbf{p}}_{\parallel}=v_{\parallel}\,\hat{\mathbf{t}},
\end{equation}
which smoothly reduces speed as the goal is approached. Lateral deviation
to the goal line is removed with a small PD term
\begin{equation}
\label{eq:land_lateral}
\dot{\mathbf{p}}_{\perp} \;=\; -k_{\perp}\, e_{\perp}\,\hat{\mathbf{n}},
\end{equation}
where \(e_{\perp}\) is the signed cross–track error and \(\hat{\mathbf{n}}\) the unit
normal of the goal line (as defined in the Travel mode).

\paragraph{Descent profile.}
When the horizontal distance drops below \(r_{\text{desc}}\) and the heading
error satisfies \(|e_{\psi}|<\psi_{\text{th}}\), the vehicle initiates a
bounded-rate vertical descent:
\begin{equation}
\label{eq:land_descent}
\dot z_{\text{ref}} \;=\; -\,\mathrm{clip}(k_z(z-z_g),\ 0,\ v_z^{\max}),
\end{equation}
with \(z_g\) the goal altitude (e.g., ground or pad height).
Equations~\eqref{eq:land_speed}–\eqref{eq:land_descent} yield a coordinated
final approach: slow, straight closing in the plane and a gentle, rate-limited
descent.

\paragraph{Abort and re-entry.}
Conditions~\eqref{eq:land_safety} are monitored throughout landing. If either
threshold is violated, we abort the descent, pause vertical motion
($\dot z_{\text{ref}}=0$), and hand control to RL for local re-planning.
The system re-enters \emph{Landing} when \eqref{eq:land_trigger} and
\eqref{eq:land_safety} are simultaneously satisfied again (after the dwell time).

\paragraph{Success and termination.}
Touchdown is declared when the following hold simultaneously for a dwell
interval \(T_{\text{settle}}\):
\begin{equation}
\label{eq:land_terminate}
\|\mathbf{p}-\mathbf{p}_g\| < r_{\text{tol}},\quad
|z-z_g|<h_{\text{tol}},\quad
\|\dot{\mathbf{p}}\|<v_{\text{tol}},\quad
|e_{\psi}|<\psi_{\text{tol}}.
\end{equation}
The episode then ends with a terminal reward. This design keeps landing
computationally light, maintains stability via hysteresis and dwell-time,
and safely defers to RL if the local environment becomes risky during the
final approach.

\section{Experimental Setup}

All experiments are run in Webots (macOS Sequoia 15.5, Apple M4 Max, 36\,GB RAM) with a DJI Mavic~2 Pro model (physics, IMU, GPS enabled). Worlds (.wbt) contain static obstacles (trees, fences, walls) and optional movers. 

\begin{figure}[t]
  \centering
  \begin{subfigure}[t]{0.45\linewidth}
    \centering
    \includegraphics[width=\linewidth]{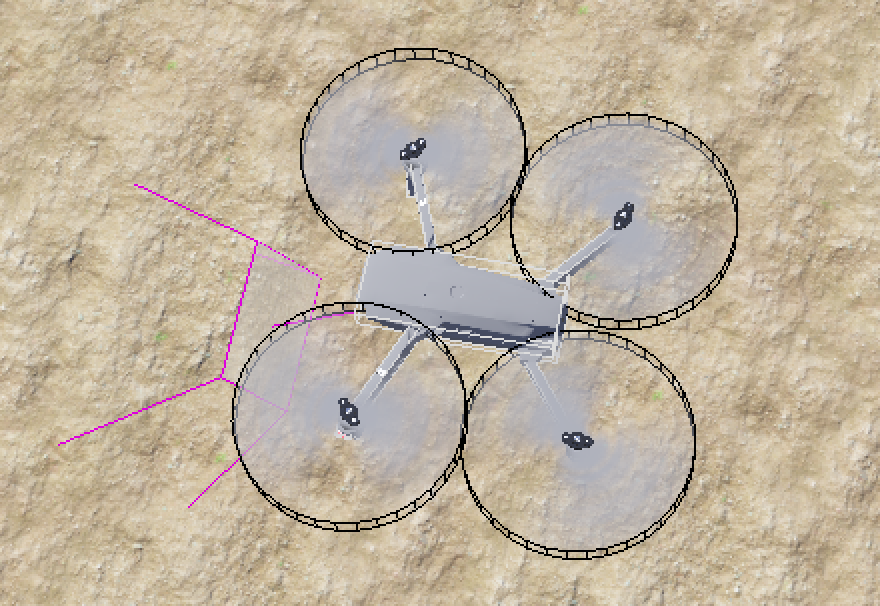} 
    \caption{Top view}
  \end{subfigure}\hfill
  \begin{subfigure}[t]{0.45\linewidth}
    \centering
    \includegraphics[width=\linewidth]{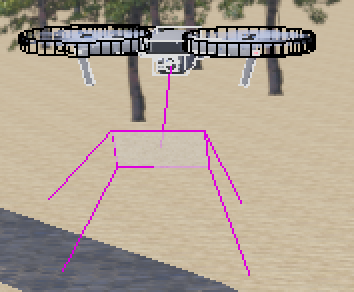}
    \caption{Front view}
  \end{subfigure}
  \caption{DJI Mavic~2 Pro in Webots.}
  \label{fig:mavic_views}
\end{figure}

\begin{figure}
  \centering
  \begin{subfigure}[t]{0.45\linewidth}
    \centering
    \includegraphics[width=\linewidth]{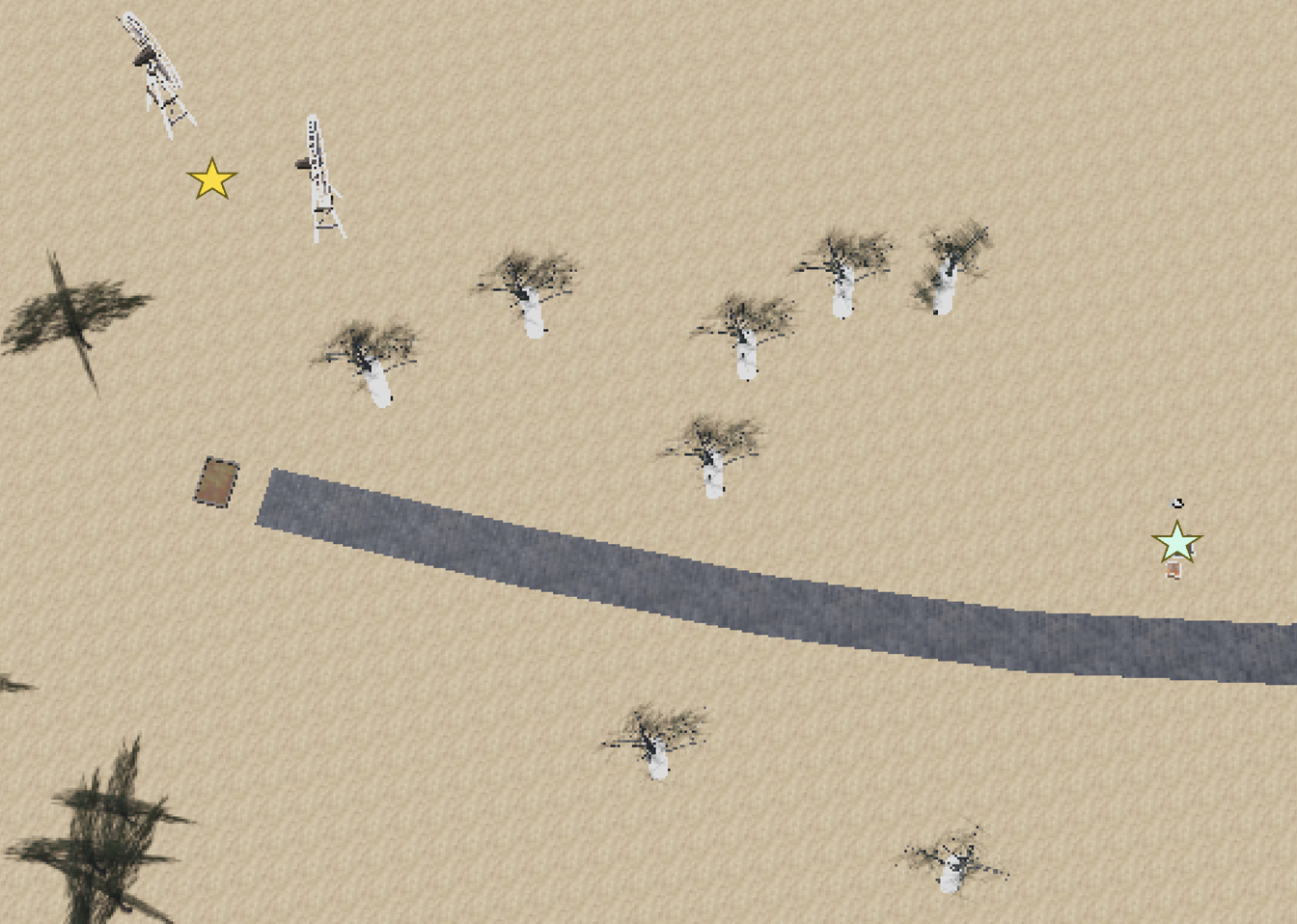}
    \caption{Trajectory~1 map.}
    \label{fig:map_traj1}
  \end{subfigure}\hfill
  \begin{subfigure}[t]{0.45\linewidth}
    \centering
    \includegraphics[width=\linewidth]{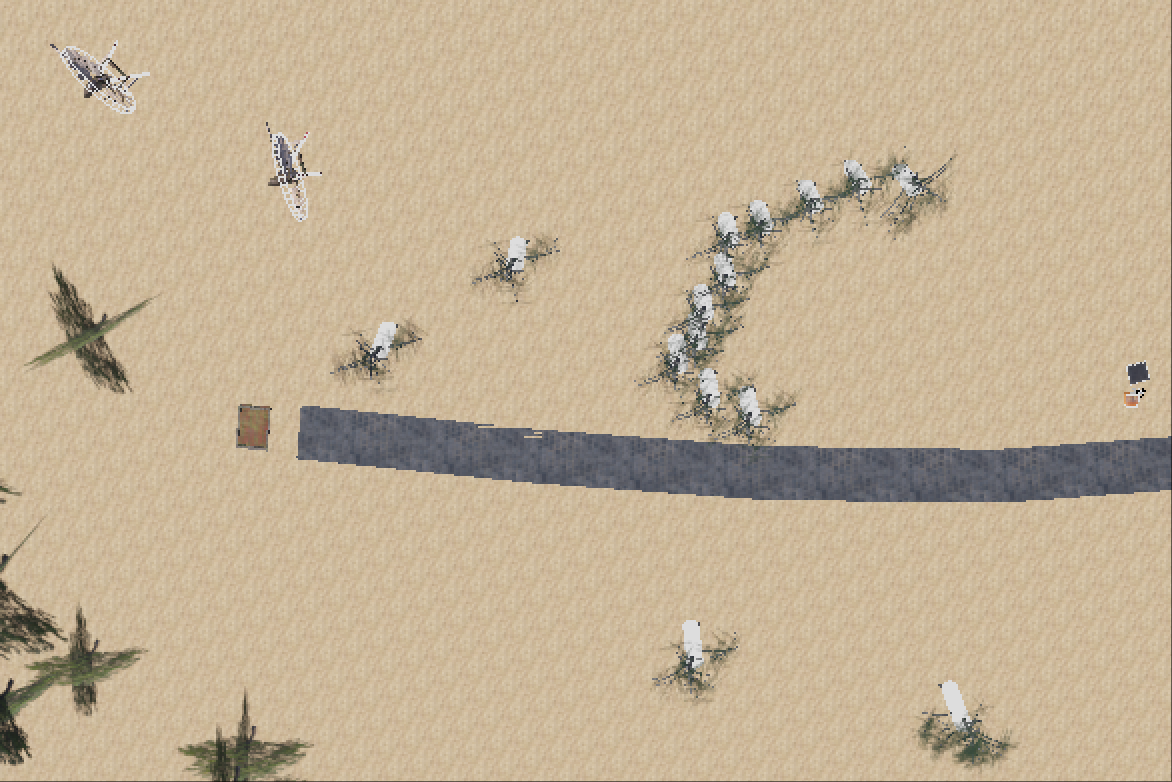}
    \caption{Trajectory~2 map.}
    \label{fig:map_traj2}
  \end{subfigure}
  \caption{Simulation maps used for evaluation (Trajectory~1 vs.\ Trajectory~2).}
  \label{fig:environments}
\end{figure}

Figure~\ref{fig:mavic_views} shows the quadrotor model. Figure~\ref{fig:environments} contrasts two evaluation maps. 
\emph{Trajectory~1} features sparse, near-uniform obstacles that form wide, well-connected corridors with largely preserved line-of-sight to the goal; the agent can follow the guidance line with only brief local corrections, so RL re-planning is rarely invoked. 
\emph{Trajectory~2} presents a denser, non-convex layout (a curved barrier plus clustered obstacles) that fragments free space into multiple homotopy classes, occludes the start–goal line, and creates narrow gaps comparable to the vehicle’s safety envelope—conditions that demand longer and more frequent RL-driven re-planning segments.

\subsection{Markov Decision Process Definition}

We model \textbf{SWIFT-Nav} as an MDP $\mathcal{M}=(\mathcal{S},\mathcal{A},P,R,\gamma)$ with continuous state/action spaces and discount $\gamma\in(0,1)$.

\paragraph{State space $\mathcal{S}$ (72-D).}
The state concatenates UAV intrinsics and a structured obstacle context:
\begin{equation}
\label{eq:state_concat}
\mathbf{s}_t=\big[\,(\mathbf{s}^{\text{intr}}_t)^\top,\ (\mathbf{c}_t)^\top\,\big]^\top \in \mathbb{R}^{72}.
\end{equation}
The 12-D intrinsic part is
\begin{equation}
\label{eq:state_intr}
\mathbf{s}^{\text{intr}}_t=\big[x,\,y,\,z,\,\text{roll},\,\text{pitch},\,\text{yaw},\,\dot r,\,\dot p,\,\dot y,\,\dot x,\,\dot y,\,\dot z\big]^\top\!\in\mathbb{R}^{12},
\end{equation}
and the 60-D obstacle context stacks up to the $20$ nearest obstacles as planar offsets and ranges
\begin{equation}
\label{eq:state_ctx}
\mathbf{c}_t=\big[(\Delta x_1,\Delta y_1,d_1),\,\ldots,\,(\Delta x_{20},\Delta y_{20},d_{20})\big]^\top\!\in\mathbb{R}^{60}.
\end{equation}
If fewer than $20$ obstacles are detected, missing triples are padded with a large range (dist$=55$) to keep a fixed dimension. This structured context provides better geometry awareness than raw proximities alone.

\paragraph{Action space $\mathcal{A}$ (10-D, waypoint interface).}
The policy outputs five body–relative 2-D waypoints that define the next local path segment. For compactness we use a matrix form and its vectorization:
\begin{equation}
\label{eq:action_mat}
\mathbf{A}_t=
\begin{bmatrix}
x_1 & y_1\\
\vdots & \vdots\\
x_5 & y_5
\end{bmatrix},\qquad
\mathbf{a}_t=\operatorname{vec}(\mathbf{A}_t)\in[-3,3]^{10}.
\end{equation}
These waypoints are tracked by a single line–following controller. The high-level waypoint interface decouples planning from low-level actuation, yielding smoother paths and stabler training than direct torque commands.

\paragraph{Reward function $R$.}
Per-step reward is the sum of time, progress, safety, and terminal terms:
\begin{equation}
\label{eq:mdp_reward}
r_t
= r_{\text{step}}
+ r_{\text{progress}}
+ r_{\text{penalty}}
+ r_{\text{goal}}
+ r_{\text{crash}}
+ r_{\text{extra}}.
\end{equation}
Here $r_{\text{step}}$ is a small time cost; $r_{\text{progress}}$ rewards reduction in goal distance; $r_{\text{penalty}}$ discourages clearance violations/unsafe proximity; $r_{\text{goal}}$ and $r_{\text{crash}}$ are terminal bonus/penalty; and $r_{\text{extra}}$ collects optional smoothness and switch-economy terms. The environment-specific coefficients are listed in Tables~\ref{tab:t1_rewards_mapped} and \ref{tab:t2_rewards_mapped}.

\begin{table}[t]\centering\scriptsize
\setlength{\tabcolsep}{5pt}\renewcommand{\arraystretch}{1.10}
\caption{Reward components and event mapping for \textbf{Trajectory~1}
(contributions to \(R=r_{\text{step}}+r_{\text{progress}}+r_{\text{penalty}}+r_{\text{goal}}+r_{\text{crash}}+r_{\text{extra}}\)).}
\label{tab:t1_rewards_mapped}
\begin{tabular}{@{}lll@{}}
\toprule
\textbf{Component \(r_{\bullet}\)} & \textbf{Event / Condition} & \textbf{Contribution} \\
\midrule
\(r_{\text{step}}\)      & every step                                   & \(-\,1.0\) \\
\(r_{\text{progress}}\)  & \(d_{\text{prev}}-d_{\text{curr}}\)          & \(+\,2.0\,(d_{\text{prev}}-d_{\text{curr}})\) \\
\(r_{\text{goal}}\)      & goal reached \((d_{\text{curr}}<2\,\text{m})\) & \(+\,200\) \\
\(r_{\text{crash}}\)     & collision (any contact)                      & \(-\,1000\) \\
\(r_{\text{penalty}}\)   & departure from guidance line                 & \(-\,50\) \\
\(r_{\text{extra}}\)     & safe obstacle bypass                         & \(+\,100\) \\
\bottomrule
\end{tabular}
\end{table}

\begin{table}[t]\centering\scriptsize
\setlength{\tabcolsep}{4pt}\renewcommand{\arraystretch}{1.08}
\caption{Reward components and event mapping for \textbf{Trajectory~2}.
(OOB = out-of-bounds.)}
\label{tab:t2_rewards_mapped}
\resizebox{\linewidth}{!}{%
\begin{tabular}{@{}lll@{}}
\toprule
\textbf{Component \(r_{\bullet}\)} & \textbf{Event / Condition} & \textbf{Contribution} \\
\midrule
\(r_{\text{step}}\)      & every step                                         & \(-\,0.5\) \\
\(r_{\text{progress}}\)  & \(d_{\text{prev}}-d_{\text{curr}}\)                & \(+\,1.5\,(d_{\text{prev}}-d_{\text{curr}})\) \\
\(r_{\text{goal}}\)      & goal reached \((d_{\text{curr}}<2\,\text{m})\)      & \(+\,3500\) \\
\(r_{\text{crash}}\)     & recoverable collision (\texttt{has\_crashed})       & \(-\,200\) \\
\(r_{\text{crash}}\)     & terminal failure (OOB or attitude violation)        & \(-\,1500\) \\
\(r_{\text{crash}}\)     & no-progress termination (displacement \(<0.5\) m over 50 steps) & \(-\,500\) \\
\(r_{\text{penalty}}\)   & proximity penalty: \(d_{\min}<5\,\text{m}\)         & \(-\,1.2\,(5-d_{\min})\) \\
\(r_{\text{penalty}}\)   & repeated RL endpoint (recently visited)             & \(-\,10\) \\
\(r_{\text{extra}}\)     & one-shot safe switch RL\(\to\)Travel                 & \(+\,500\) \\
\(r_{\text{extra}}\)     & exit from obstacle zone \((d_{\min}>5\,\text{m})\)   & \(+\,50\) \\
\(r_{\text{extra}}\)     & RL endpoint novelty                                 & \(+\,20\) \\
\bottomrule
\end{tabular}%
}
\end{table}

\subsection{Prioritized Experience Replay (PER)}
We use a 50k-capacity buffer (batch 128). New samples receive $p_{\max}$ to be sampled at least once; sampling probability is $P(i)\propto p_i^{\alpha}$ with $\alpha=0.6$. Importance-sampling weights $w_i$ correct bias, with $\beta$ annealed from 0.4 to 1.0. After each critic update, TD-errors refresh priorities. The same PER setup is used in both scenes; the overall TD3+PER training loop is summarized in Algorithm~\ref{alg:td3_per}.

\subsection{Epsilon-Greedy Exploration in Waypoint Space}
Exploration is injected at the waypoint-parameter level (not at motor commands). Upon entering the RL mode, with \emph{probability} \(\epsilon\) we replace the actor’s output with a vector sampled uniformly from \([-1,1]^d\) (the first two \emph{dimensions} are scaled to increase heading/curvature diversity); otherwise, the deterministic action is used. The parameters are decoded into (i) detour side, (ii) curvature-blending coefficient, and (iii) waypoint spacing, after which a locally feasible path is synthesized. If a wide free-space sector is detected, an \emph{escape line} is appended, and goal/guard hints assist re-alignment. The exploration rate \(\epsilon\) decays from \(1.0\) to \(\epsilon_{\min}=0.05\) with per-episode factor \(\lambda=0.995\).

\section{Results}

To assess the effectiveness of our approach, we run three experiments: (i) a \emph{vanilla TD3} baseline on \textbf{Trajectory~1}; (ii) \textbf{SWIFT-Nav} on \textbf{Trajectory~1}; and (iii) \textbf{SWIFT-Nav} on \textbf{Trajectory~2}. Trajectory~1 and Trajectory~2 are two maps that differ primarily in obstacle density and corridor connectivity (Trajectory~2 has denser, more fragmented free space). For each setting we present the resulting navigation trajectories to qualitatively assess path-planning behavior. In addition, for SWIFT-Nav we plot 15-episode moving-average curves of episode length (steps) and cumulative return over the course of training \emph{until 100 successful episodes have been achieved} on each map; thus, the total number of episodes differs by map and configuration.

For the baseline, we attempted the same “100 successes” protocol on Trajectory~1 under an equal training budget, but \emph{vanilla TD3} exhibited non-convergent, high-variance learning with frequent terminations; accordingly, we include representative trajectories and qualitative observations rather than smoothed learning curves. Given this instability under identical settings, we did not extend the baseline to Trajectory~2. Taken together, the trajectory visualizations and SWIFT-Nav’s learning curves provide a comprehensive view of stability and efficiency across maps with differing obstacle layouts.

\subsection{Trajectories and Learning Progress}

\begin{figure}[t]
  \centering
  \includegraphics[width=1\linewidth]{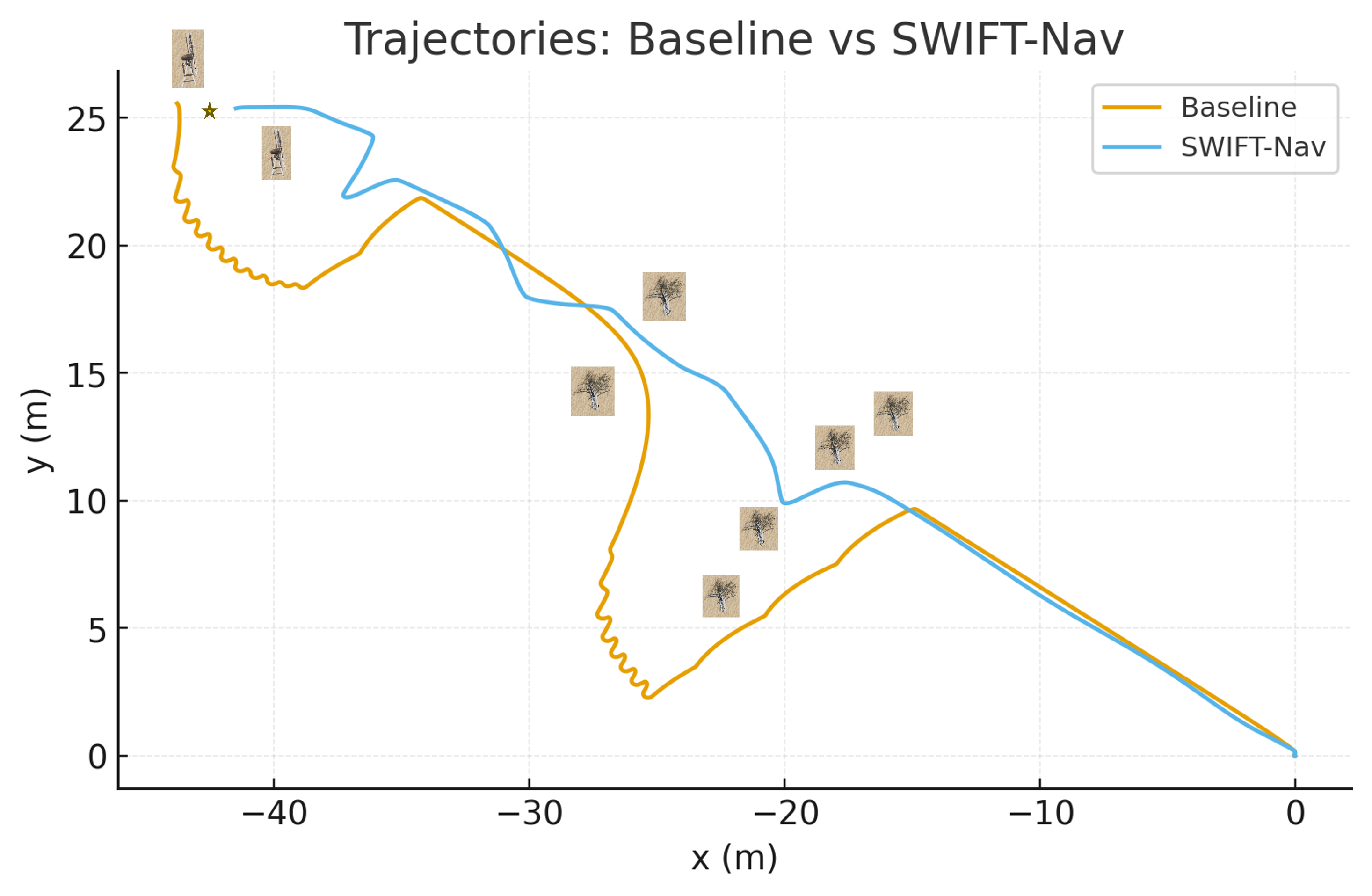}
  \caption{Comparing trajectories on \textbf{Trajectory~1}: SWIFT-Nav produces a shorter, smoother route with fewer detours, whereas the baseline path deviates more around clutter.}
  \label{fig:traj1_compare}
\end{figure}

\begin{figure*}[t]
  \centering
  \begin{subfigure}[t]{0.48\textwidth}
    \centering
    \includegraphics[width=\linewidth]{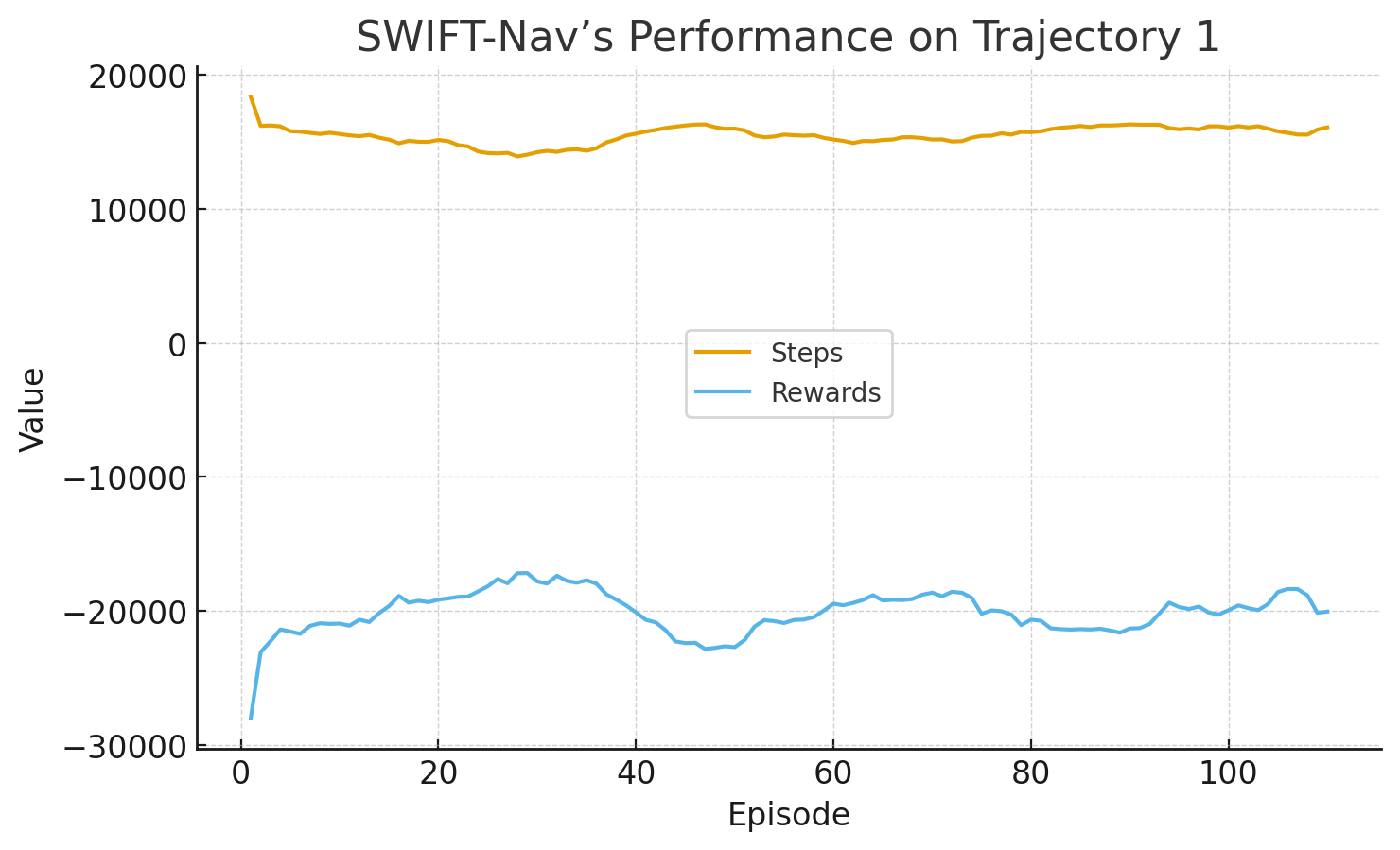}
    \caption{SWIFT-Nav's Performance on Trajectory~1}
    \label{fig:traj1_perf}
  \end{subfigure}
  \hfill
  \begin{subfigure}[t]{0.48\textwidth}
    \centering
    \includegraphics[width=\linewidth]{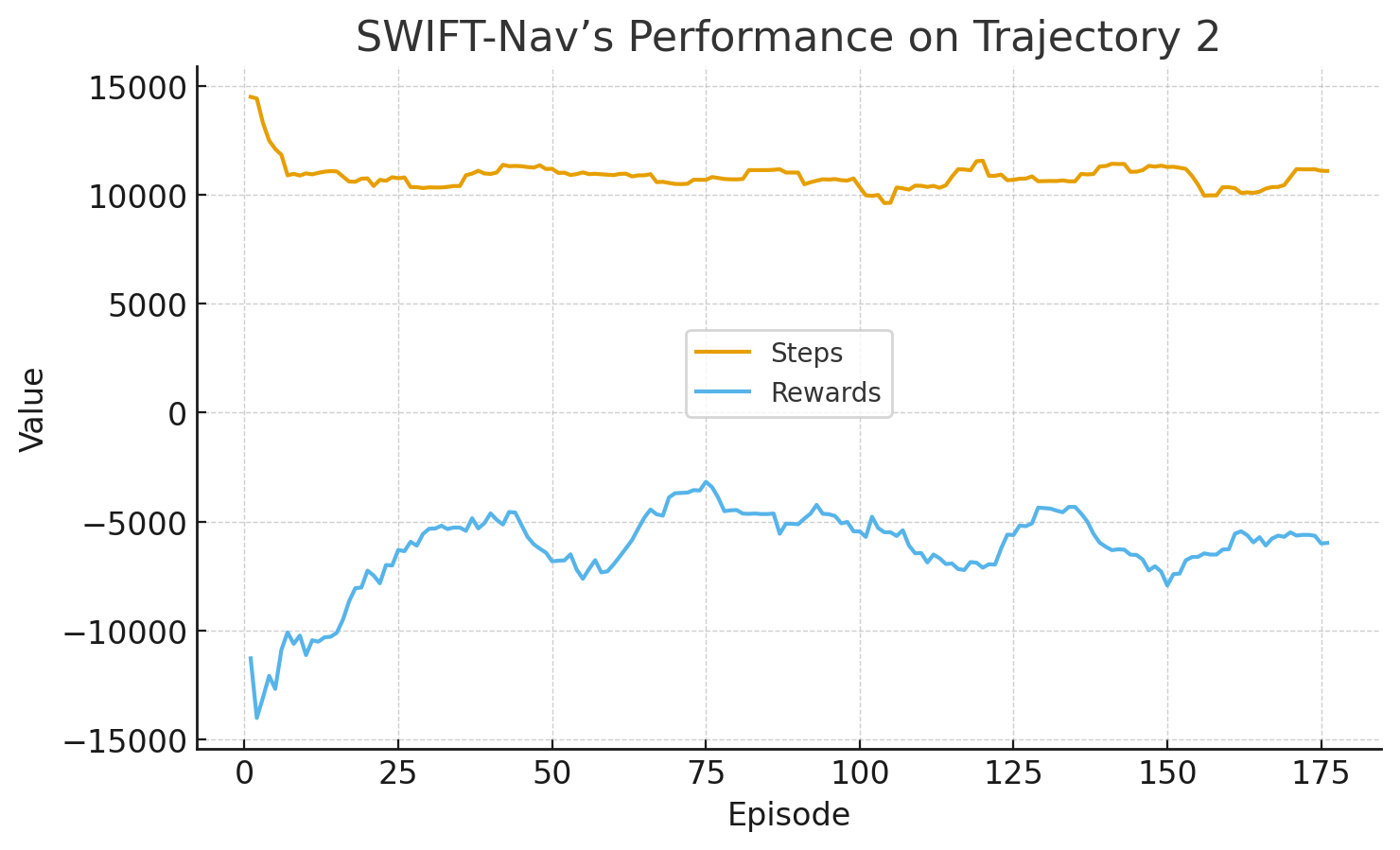}
    \caption{SWIFT-Nav's Performance on Trajectory~2}
    \label{fig:traj2_perf}
  \end{subfigure}

  \caption{15-episode moving averages of episode length and cumulative return on the two Trajectory environments.}
  \label{fig:perf_both}
\end{figure*}

We unify the qualitative path evidence and quantitative learning trends here. In Figure~\ref{fig:traj1_compare}, the differences between the baseline and our method are visually apparent across two benchmark routes (Trajectory~1 and Trajectory~2); complementing this, the 15-episode moving-average curves (Figures~\ref{fig:traj1_perf}–\ref{fig:traj2_perf}) corroborate the same convergence behavior.

\paragraph{Trajectories.}
\textbf{Baseline (vanilla TD3) on Trajectory~1} (Figure~\ref{fig:traj1_compare}): takes a noticeably longer route with multiple detours and local oscillations, suggesting an unstable policy and weak convergence under the same budget.

\textbf{SWIFT-Nav on Trajectory~1} (Figure~\ref{fig:traj1_compare}): follows a short, near-straight path with smooth turns and no visible oscillation, indicating faster learning and markedly better stability.

\begin{figure}[t]
  \centering
  \begin{subfigure}[t]{1\linewidth}
    \centering
    \includegraphics[width=\linewidth]{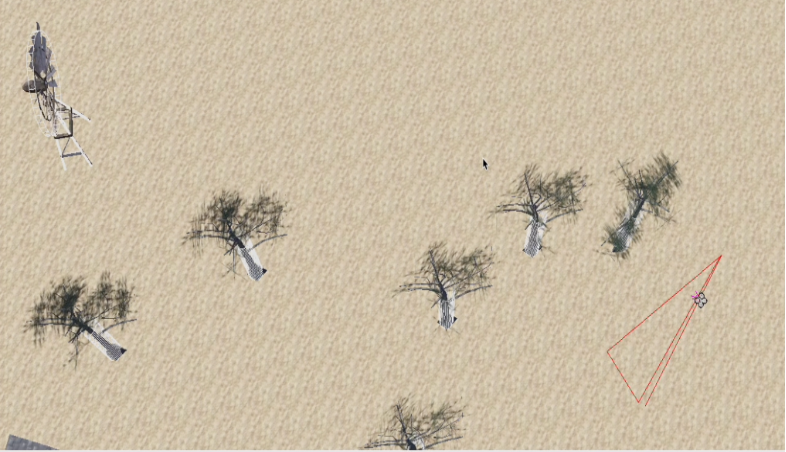}
    \caption{Baseline TD3 — local, reactive detours (triangle shows local planning footprint).}
    \label{fig:traditional}
  \end{subfigure}\par\vspace{0.6em}
  \begin{subfigure}[t]{1\linewidth}
    \centering
    \includegraphics[width=\linewidth]{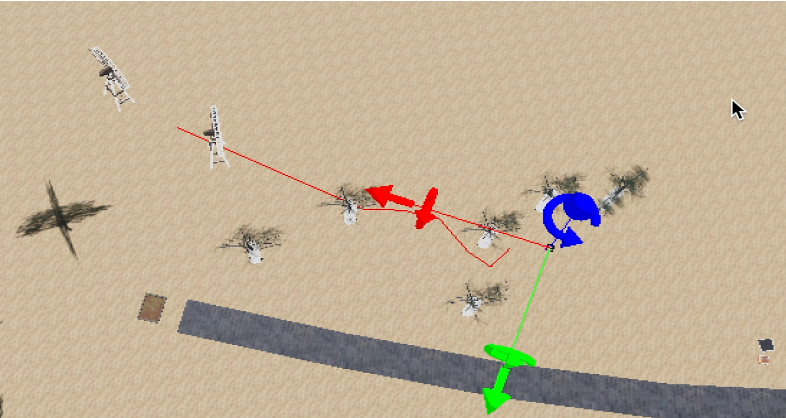}
    \caption{\textbf{SWIFT-Nav} — goal-directed, smooth trajectory through clutter.}
    \label{fig:improve}
  \end{subfigure}
  \caption{Local-only vs.\ goal-directed planning (why the paths differ).  
  (a) The baseline focuses on immediate obstacle avoidance and tends to skirt the field.  
  (b) \textbf{SWIFT-Nav} integrates local safety with the global objective, yielding a shorter, coherent route.}
  \label{fig:td3_comparison}
\end{figure}

Figure~\ref{fig:td3_comparison} clarifies the mechanism behind the above contrast.  
Panel~(a) shows the \emph{baseline TD3}: the red triangular wedge visualizes its local planning footprint, so the policy reacts only inside the nearby obstacle patch—skirting trees and producing myopic, reactive detours.  
Panel~(b) shows \emph{SWIFT-Nav}: geometry-space exploration perturbs the five proposed waypoints (rather than issuing motor-level random actions), and the trajectory checker biases proposals toward goal-aligned, collision-free polylines. The resulting plan remains smooth and globally directed while still clearing local obstacles.

\begin{figure}[t]
  \centering
  \includegraphics[width=1\linewidth]{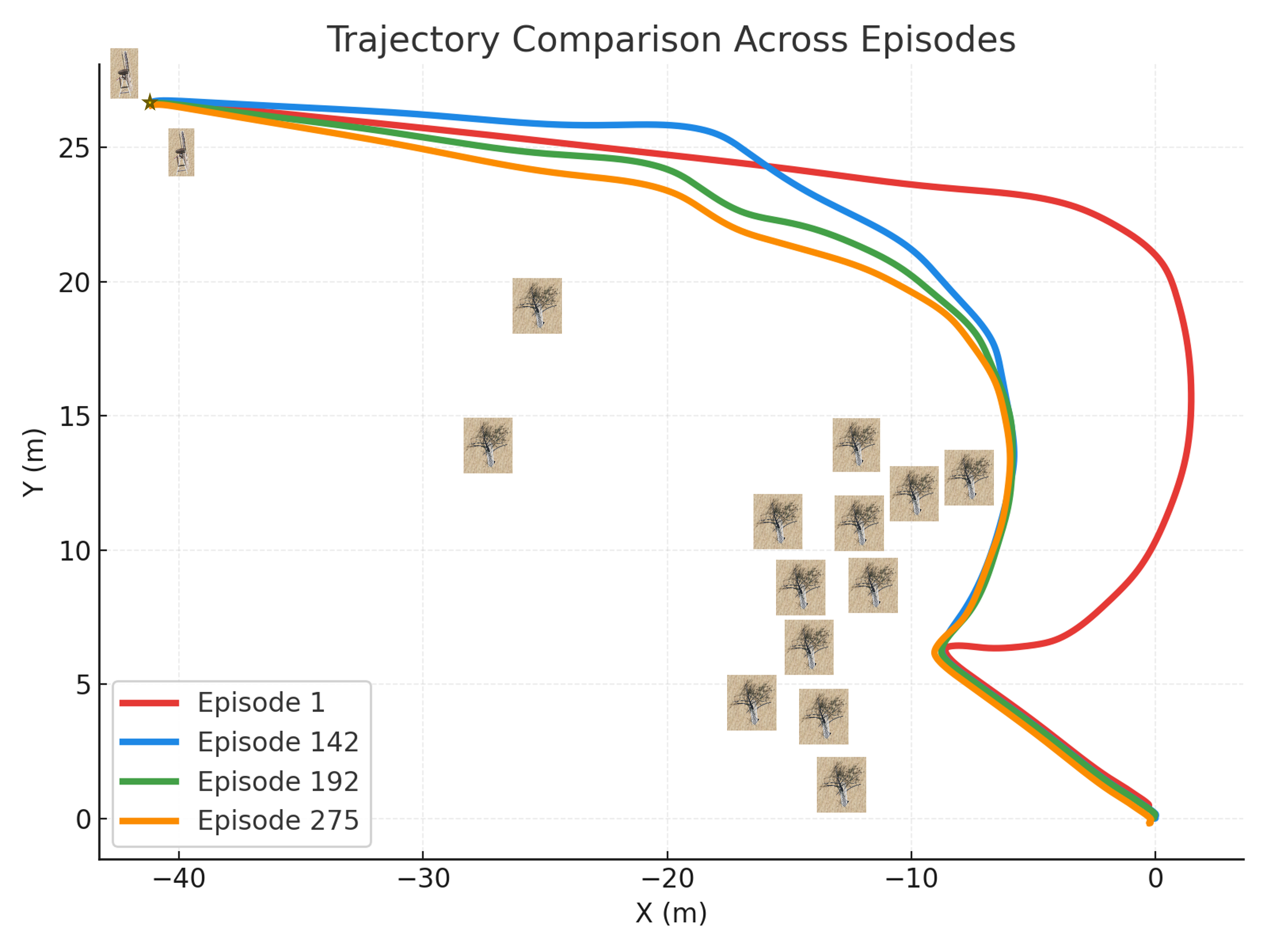}
  \caption{Trajectories across episodes on \textbf{Trajectory~2} detours shrink over training and paths become more direct and smooth.}
  \label{fig:complex_together}
\end{figure}

\paragraph{Trajectory~2 trend.}
For Trajectory~2 specifically, Figure~\ref{fig:complex_together} shows a clear training trend: early episodes exhibit large detours; by later episodes (e.g., 1/142/192/275) paths become progressively shorter and more direct, indicating that the policy has internalized both global goal seeking and local obstacle avoidance, leading to better spatial efficiency.

\paragraph{Steps and Rewards.}
Mirroring the trajectory evidence, the smoothed training curves further quantify convergence: Figure.~\ref{fig:perf_both}\subref{fig:traj1_perf} reports performance on Trajectory~1, and Figure.~\ref{fig:perf_both}\subref{fig:traj2_perf} on Trajectory~2. For the baseline, \emph{vanilla TD3} failed to converge on Trajectory~1 under the same training budget and was therefore not extended to Trajectory~2; accordingly, we omit baseline learning curves.

For \textbf{Trajectory~1}, the training converges quickly and remains steady: within a few dozen episodes, the 15-episode moving averages of steps and return collapse into a tight band, consistent with the short, clean paths. 
For \textbf{Trajectory~2}, despite the denser layout, the smoothed return rises substantially (on the order of \(10^{4}\)) while the step count stays bounded and trends downward; brief mid-training ripples from exploration fade as the policy stabilizes. 
Across both routes, the desired pattern—steps \(\downarrow\) and reward \(\uparrow\) with shrinking variance—holds consistently. (Vanilla TD3 did not converge under the same budget, so baseline curves are omitted.)

The joint trend—steps decreasing while returns increasing with reduced variance—directly reflects our stability-aware arbitration. 
Hysteresis, debounce, and dwell filter transient mode triggers; the LOS guard prevents premature exits; and the trajectory checker regularizes updates with safe, smooth proposals. 
The ablation results in Fig.~\ref{fig:ablation} further corroborate this mechanism.

\subsection{Ablation Studies}
\paragraph{Ablation setup.}
We evaluate three configurations with identical seeds and evaluation protocol:  
(a) \textbf{Baseline} — stability logic and trajectory checker both enabled;  
(b) \textbf{No-Stability} — hysteresis, debounce, and dwell time disabled;  
(c) \textbf{No-Checker} — trajectory checker bypassed while keeping stability logic active.  

\begin{figure*}[t]
  \centering
  \begin{subfigure}[t]{0.48\textwidth}
    \includegraphics[width=\linewidth]{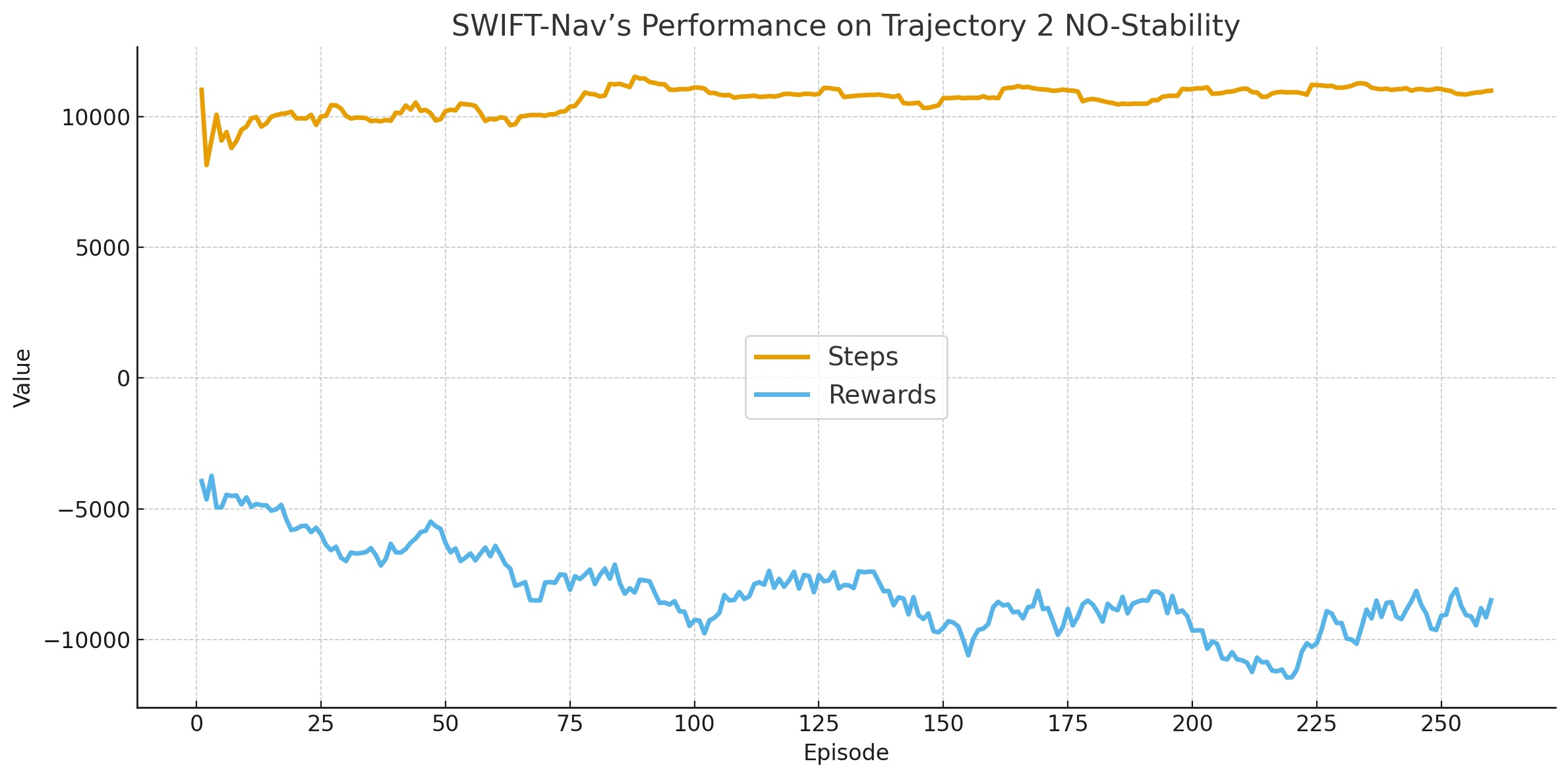}
    \caption{Ablation: No-Stability — hysteresis/debounce/dwell off.}
    \label{fig:ablation_nostable}
  \end{subfigure}
  \hfill
  \begin{subfigure}[t]{0.48\textwidth}
    \includegraphics[width=\linewidth]{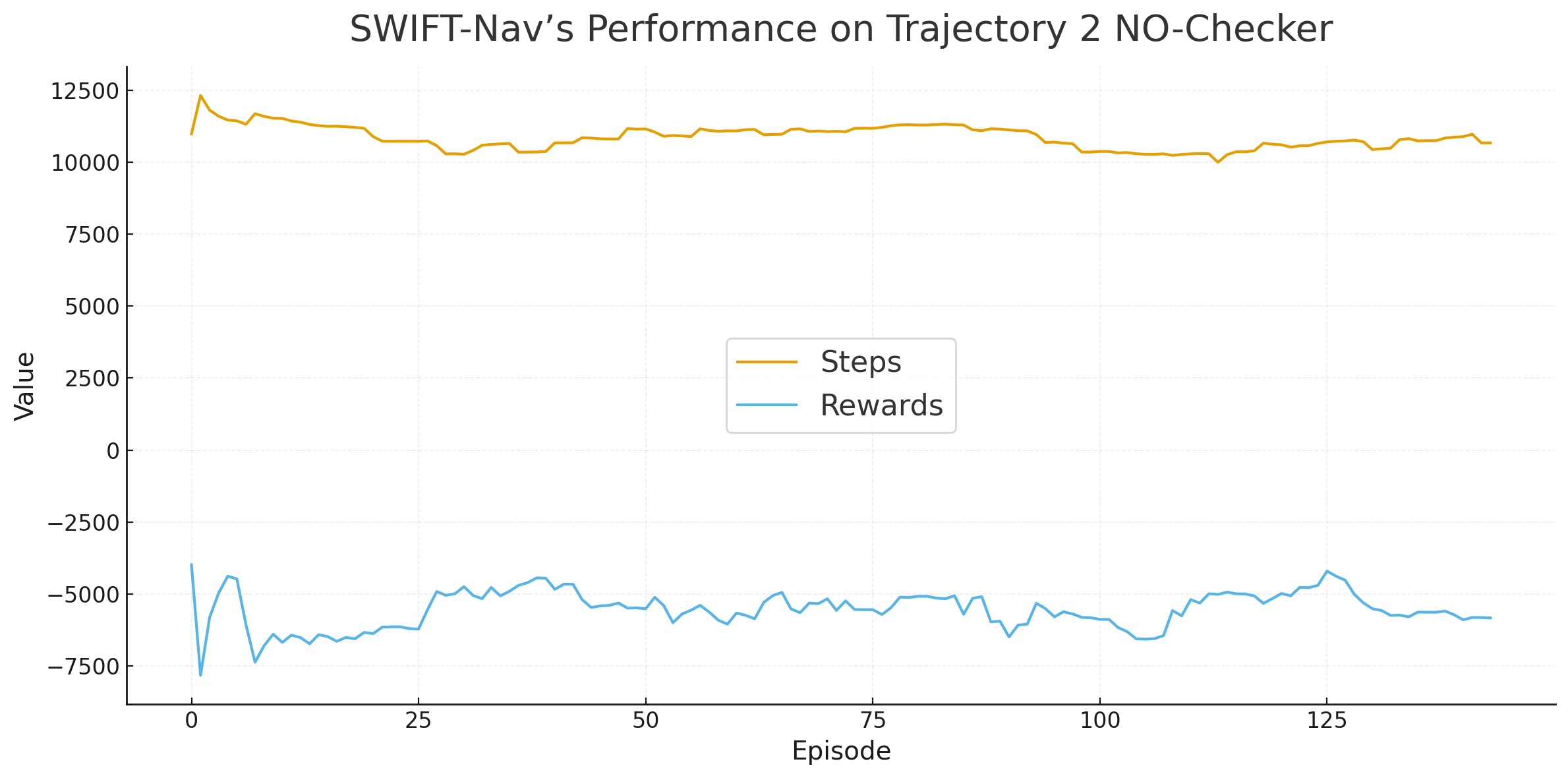}
    \caption{Ablation: No-Checker — trajectory checker off.}
    \label{fig:ablation_nochecker}
  \end{subfigure}
  \caption{Ablation study comparing the effect of disabling stability logic and the trajectory checker on SWIFT-Nav’s learning performance.}
  \label{fig:ablation}
\end{figure*}

\paragraph{Results and analysis.}
Across all runs, the \textbf{Baseline} (Fig.~\ref{fig:traj2_perf}) shows a desirable trend: the smoothed step count remains bounded and slightly decreases over training, while the cumulative return steadily increases and stabilizes—indicating efficient exploration and convergence toward a stable, high-performing policy.  
In contrast, \textbf{No-Stability} (Fig.~\ref{fig:ablation_nostable}) clearly degrades both metrics: the return curve exhibits a downward trend with large oscillations, and the step count fluctuates heavily. This confirms that removing hysteresis, debounce, and dwell time causes unstable switching between modes, leading to inconsistent performance.  
Meanwhile, \textbf{No-Checker} (Fig.~\ref{fig:ablation_nochecker}) shows almost flat return and step curves, implying that learning stagnates early. Without the lightweight geometric validator, the policy fails to refine its waypoint proposals—producing safe but overly conservative trajectories with limited improvement over time.  
Overall, these results demonstrate that both the stability logic and the trajectory checker are essential: stability logic ensures smooth and consistent arbitration, while the checker enforces geometric validity and contributes to efficiency gains during long-horizon training.

\subsection{Discussion}

SWIFT-Nav lifts policy learning from motor commands to \emph{waypoints}. The TD3 actor proposes short polylines, while a lightweight line-following controller executes them. This decoupling removes actuator lag from credit assignment, yielding steadier closed-loop behavior: the controller focuses on tracking, and TD3 decides \emph{where} to go. Mode usage is deliberate—RL is engaged only near hazards; elsewhere a low computing travel mode advances along a dynamic main line. Stability is enforced with hysteresis, debounce, and minimum dwell time, and switches are gated by a fuzzy global safety score computed from a single radar stream. Prioritized replay and a lightweight waypoint checker (guard/hint points, escape segment when wide free space is seen) further accelerate learning. A task-aligned reward mixes goal progress, clearance shaping, and switch economy, keeping the stack single-sensor and deployable.

Trajectories in Figure~\ref{fig:traj1_compare} highlight the contrast on \textbf{Trajectory~1}: the vanilla TD3 baseline traces a long, wiggly path with unnecessary detours, whereas \textbf{SWIFT\mbox{-}Nav} produces a short, smooth route with only a few, sustained mode switches. The learning curves in Fig.~\ref{fig:perf_both} corroborate this behavior: in \subref{fig:traj1_perf} (Trajectory~1) the 15-episode moving averages rapidly collapse into a tight band (low variance, stable policy), and in \subref{fig:traj2_perf} (Trajectory~2) the smoothed return rises by roughly \(10^{4}\) while step counts remain controlled and trend downward, with only brief exploratory ripples.

Reasoning at the waypoint level—combined with stability-aware arbitration and PER-driven efficiency—delivers short, smooth trajectories, rare non-oscillatory switches, and the characteristic ``steps decrease and reward increase'' signature, without heavy multi-sensor perception or high compute.

\section{Conclusions}

We introduced \textbf{SWIFT-Nav}, a stability-aware, waypoint-level TD3 framework for UAV navigation in cluttered scenes. 
By decoupling \emph{where to go} (waypoint planning) from \emph{how to get there} (a shared line-following controller) and engaging RL only when risk is detected, 
SWIFT-Nav delivers fast, stable closed-loop behavior under a single range sensor. 
A prioritized-replay pipeline with directional, decaying exploration in waypoint space and a lightweight trajectory checker regularizes the update stream without heavy perception or compute.

\textbf{Empirical takeaways.} 
Across two maps with distinct obstacle layouts, SWIFT-Nav consistently produced shorter and smoother trajectories than a vanilla TD3 baseline and exhibited markedly steadier learning dynamics: 
on Trajectory~1, 15-episode moving averages of return and step count collapse into a narrow band within tens of episodes; 
on Trajectory~2, cumulative return rises while steps trend downward, indicating improved efficiency without sacrificing stability. 
The qualitative gap matches the mechanism: the baseline reacts within a local planning footprint, whereas our geometry-space exploration and checker bias proposals toward the global goal while preserving clearance.

\textbf{Practical significance.} 
The architecture maintains real-time responsiveness and a low compute footprint, 
and we provide a \emph{Webots-based simulation pipeline that runs natively on Apple Silicon (M-series) Macs}, 
filling a tooling gap left by mainstream simulators that target Windows/Linux. 
This lowers the barrier to reproduction and iteration on commodity macOS laptops.

\textbf{Limitations and future work.} 
Our study is simulation-only and uses a single range-sensor stack; we also report single-seed curves due to time and budget constraints. 
Future work will include hardware flights to assess sensing delay and noise, 
multi-seed runs with confidence bands, and finer ablations (PER, $\varepsilon$-greedy schedule, checker components) to quantify each contribution. 
We also plan to evaluate on additional maps with varying clutter levels and release our Apple-Silicon Webots setup and trained policies to encourage reproducibility and community benchmarking.


\begin{thebibliography}{}
\bibitem[\protect\citeauthoryear{Muchiri and Kimathi}{2022}]{mu22}
G.~N. Muchiri and S. Kimathi.
\newblock A review of applications and potential applications of {UAV}.
\newblock In \emph{Proceedings of the Sustainable Research and Innovation Conference},
pages 280--283, 2022.

\bibitem[\protect\citeauthoryear{Yin \bgroup \em et al.\egroup}{2024}]{yin24}
Y.~Yin, Z.~Wang, L.~Zheng, Q.~Su, and Y.~Guo.
\newblock Autonomous {UAV} navigation with adaptive control based on deep reinforcement learning.
\newblock \emph{Electronics}, 13(13):2432, 2024.


\bibitem[\protect\citeauthoryear{Fagundes-Junior \bgroup \em et al.\egroup}{2024}]{fag24}
L.~A.~Fagundes-Junior, K.~B.~de~Carvalho, R.~S.~Ferreira, and A.~S.~Brand{\~a}o.
\newblock Machine learning for unmanned aerial vehicles navigation: An overview.
\newblock \emph{SN Computer Science}, 5(2):256, 2024.

\bibitem[\protect\citeauthoryear{Zhao \bgroup \em et al.\egroup}{2024}]{zha24}
Y.~Zhao, J.~Zhang, and C.~Zhang.
\newblock Deep-learning based autonomous-exploration for UAV navigation.
\newblock \emph{Knowledge-Based Systems}, 297:111925, 2024.

\bibitem[\protect\citeauthoryear{Hussain \bgroup \em et al.\egroup}{2024}]{hus24}
A.~Hussain, S.~Li, T.~Hussain, X.~Lin, F.~Ali, and A.~A.~AlZubi.
\newblock Computing Challenges of UAV Networks: A Comprehensive Survey.
\newblock \emph{Computers, Materials \& Continua}, 81(2), 2024.

\bibitem[\protect\citeauthoryear{Ko \bgroup \em et al.\egroup}{2025}]{ko25}
T.~Ko, J.~Park, S.~Choi, and J.~Shim.
\newblock Autonomous Flight of UAV in Complex Multi-Obstacle Environment Using Data-Driven and Vision-Based Deep Reinforcement Learning and AirSim.
\newblock In \emph{AIAA Aviation Forum and ASCEND}, paper 3686, 2025.

\bibitem[\protect\citeauthoryear{Guo \bgroup \em et al.\egroup}{2024}]{guo24}
C.~Guo, C.~Sheng, H.~An, H.~Wang, X.~Lu, and J.~Nie.
\newblock Improved DQN for Path Planning with Classified High-value Prioritized Experience Replay.
\newblock In \emph{Proceedings of the 2024 China Automation Congress (CAC)}, pages 3163--3167. IEEE, 2024.

\bibitem[\protect\citeauthoryear{Mill{\'a}n \bgroup \em et al.\egroup}{2019}]{millan19}
C.~Mill{\'a}n, B.~J.~T. Fernandes, and F.~Cruz.
\newblock Human feedback in continuous actor--critic reinforcement learning.
\newblock In \emph{Proceedings of the European Symposium on Artificial Neural Networks, Computational Intelligence and Machine Learning (ESANN)}, 2019.

\bibitem[\protect\citeauthoryear{He et al.}{2020}]{he20}
Lei He, Nabil Aouf, James F. Whidborne, and Bifeng Song.
\newblock Deep reinforcement learning based local planner for UAV obstacle avoidance using demonstration data.
\newblock {\em arXiv preprint arXiv:2008.02521}, 2020.

\bibitem[\protect\citeauthoryear{Luo et al.}{2024}]{Luo24}
Xuqiong Luo, Qiyuan Wang, Hongfang Gong, and Chao Tang.
UAV Path Planning Based on the Average TD3 Algorithm with Prioritized Experience Replay.
\textit{IEEE Access}, 12:38017--38029, 2024.

\bibitem[\protect\citeauthoryear{Li et al.}{2024}]{Li24}
Peng Li, Donghui Chen, Yuchen Wang, Lanyong Zhang, and Shiquan Zhao.
Path planning of mobile robot based on improved TD3 algorithm in dynamic environment.
\textit{Heliyon}, 10(11), 2024.

\bibitem[\protect\citeauthoryear{Xia, Mantegh, and Xie}{2024}]{Xia24}
Bingze Xia, Iraj Mantegh, and Wen-Fang Xie.
\newblock Hybrid framework for UAV motion planning and obstacle avoidance:
Integrating deep reinforcement learning with fuzzy logic.
\newblock In \emph{Proceedings of the 10th International Conference on Control, Decision and Information Technologies (CoDIT)}, pages 2662--2669. IEEE, 2024.

\bibitem[\protect\citeauthoryear{Liu, McCane, and Mills}{2024}]{Liu24}
Juncheng Liu, Brendan McCane, and Steven Mills.
\newblock Learning to explore by reinforcement over high-level options.
\newblock \emph{Machine Vision and Applications}, 35(1):6, 2024.
\newblock Springer.

\end{thebibliography}

\end{document}